%% file: main.tex
\documentclass[runningheads]{llncs}
\usepackage[utf8]{inputenc}
\usepackage{graphicx}
\usepackage{subcaption}
\usepackage[table,xcdraw]{xcolor}
\usepackage{url}
\usepackage{float}
\usepackage{amsmath}
\usepackage{multirow}
\usepackage{comment}
\usepackage{multicol}
\usepackage{wrapfig}
\usepackage{multirow}
\usepackage[table,xcdraw]{xcolor}
\newcommand{\fref}[1]{Figure \ref{#1}}
\newcolumntype{L}[1]{>{\raggedright\let\newline\\\arraybackslash\hspace{0pt}}m{#1}}
\newcolumntype{C}[1]{>{\centering\let\newline\\\arraybackslash\hspace{0pt}}m{#1}}
\newcolumntype{R}[1]{>{\raggedleft\let\newline\\\arraybackslash\hspace{0pt}}m{#1}}
\begin{document}
\title{Planning the path with Reinforcement Learning: Optimal Robot Motion Planning in RoboCup Small Size League Environments}
%
\titlerunning{Planning the path with Reinforcement Learning}
%
\author{
Mateus G. Machado,
Jo\~ao G. Melo,
Cleber Zanchettin,
Pedro H. M. Braga,
Pedro V. Cunha,
Edna N. S. Barros \and
Hansenclever F. Bassani}
%
%
\institute{Centro de Informática, Universidade Federal de Pernambuco. \\
Av. Prof. Moraes Rego, 1235 - Cidade Universitária, Recife - Pernambuco, Brazil.
\email{\{mgm4, jgocm, cz, phmb4, pvc3, ensb, hfb\}@cin.ufpe.br}}

\authorrunning{Machado et. al.}
%
\maketitle              

\input{content/abstract.tex}

\input{content/intro}
\input{content/related}
\input{content/approach}
\input{content/stability}
\input{content/results}

\input{content/discussion}
\input{content/conclusion}


%
%
%
\bibliographystyle{splncs04}

\bibliography{references}

\end{document}

%% file: content/abstract.tex
\begin{abstract}
This work investigates the potential of Reinforcement Learning (RL) to tackle robot motion planning challenges in the dynamic RoboCup Small Size League (SSL). Using a heuristic control approach, we evaluate RL's effectiveness in obstacle-free and single-obstacle path-planning environments. Ablation studies reveal significant performance improvements. Our method achieved a \textbf{60\% time gain} in obstacle-free environments compared to baseline algorithms. Additionally, our findings demonstrated \textbf{dynamic obstacle avoidance capabilities}, adeptly navigating around moving blocks. These findings highlight the potential of RL to enhance robot motion planning in the challenging and unpredictable SSL environment.
\keywords{Reinforcement Learning, Motion Planning, Robot Control}
\end{abstract}

%% file: content/intro.tex
\section{Introduction} \label{sec:intro}
The RoboCup Small Size League (SSL) is a robot soccer competition that focuses on the problem of multi-robot cooperation and control in a highly dynamic environment \cite{ssl-rules}. This challenging game requires controlling a team of up to 11 robots to work together to perform passes, dribbles, and shots on goal in a coordinated and effective way. The effectiveness of these actions hinges on precise path-planning tailored to the distinct motion models of each robot \cite{pathPlanningOverview}.
\begin{figure}[!ht]
    \includegraphics[width=\linewidth]{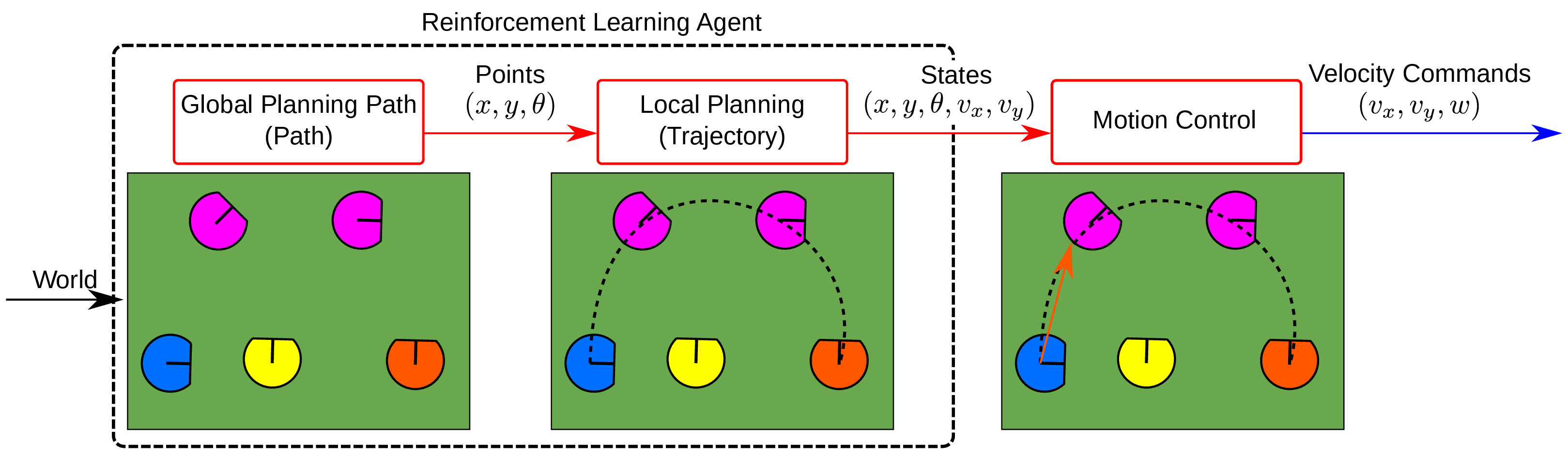}
    \caption{General workflow for robot motion planning and control, divided into Global planning, local planning, and motion control. The first two stages plan the points and velocities the agent should follow, while the last controls the robot. This Work focuses on the first two modules of this workflow.}
    \label{fig:general-workflow}
\end{figure}

Figure \ref{fig:general-workflow} represents the general workflow for planning a robot's motion within the SSL environment. Given a target position and orientation in the environment (depicted in orange), the process unfolds in three key steps. Initially, global planning calculates a geometric path, represented by a set of coordinates in the configuration space, guiding the robot toward the desired target. Subsequently, the local planner generates a sequence of desired states, comprising positions and velocities, forming a trajectory aligning with the previously computed path. Finally, the motion control module determines the requisite control inputs for the robot to transition from its current state to the next desired state.

Reinforcement Learning (RL) has emerged as a promising approach for robotic control tasks\cite{ddpg,rsoccer}. RL offers a paradigm where agents learn optimal strategies through interaction with their environment. This learning paradigm finds a natural application in the SSL, as it allows agents to adapt and improve their decision-making processes in response to the dynamic and unpredictable nature of the game  \cite{sutton}. 

We discern a hierarchical structure by adapting the motion planning workflow to RL learning environments. While the motion control module directly governs the robot's actions, the global and local planning modules serve as intermediary sub-goals to be achieved. We refer to the motion control task as \textbf{goToPoint}, with global and local planning denoted as \textbf{path-planning}.

Building upon this foundation, we propose a model-free path-planning methodology, leveraging goal-conditioned policies that take the goal state as input \cite{kaelbling1993learning,pong2018temporal}, and yield sub-goals to be executed by an omnidirectional motion control \cite{siegwart2011introduction}.
Sim2Real is a subfield of Machine Learning focused on mitigate the gap between simulated and real-world environments \cite{10242366,Gao_2022_CVPR}. While RL environments traditionally model the entire motion planning system, integrating a Sim2Real module becomes imperative to transfer learned models to real-world settings. Our methodology circumvents this challenge, as the path-planning model remains agnostic to the training environment, whether simulated or real.

Our primary contribution lies in a methodology for employing Reinforcement Learning in the path-planning task, creating a model that seamlessly transitions from simulation to reality \cite{pathPlanningOverview,motionPlanninReview}. We evaluate the efficacy of the Soft Actor-Critic (SAC) \cite{sac} algorithm across both baseline and proposed learning environments, shedding light on discrepancies inherent in the baseline's reliance on goToPoint to address the path-planning task. Additionally, we introduce two state-of-the-art methods to mitigate action instability and craft intuitive trajectories. Finally, through empirical validation, we demonstrate the readiness of our model for real-world deployment.


%% file: content/related.tex
\section{Related Work}\label{sec:related}

Traditionally, motion planning systems are conceptualized as single complex agents aimed at solving the entire system \cite{eysenbach2019search,fang2018dher,nair2018visual}. However, a hierarchical understanding of the system \cite{FeudalRL,LearningMultiLevelHierarchies} delineates a manager-worker structure, with the path-planning task delegated to a manager agent and goToPoint executed by a worker agent. In our approach, we base the environment development on the managerial treatment of the environment in existing works.

In Reinfocement Learning, the reward function quantifies the impact of an action on the environment and serves as a crucial component to guide the agent through the learning process \cite{sutton}. Our baseline work \cite{ze} endeavors to solve the path-planning task for robots in the Small Size League (SSL) of RoboCup \cite{ssl-rules}. As elucidated in subsequent sections, \cite{ze} adopts a reward system emphasizing direct impact actions, culminating in an inherently unstable goToPoint agent.

The challenge of ensuring the robustness of RL algorithms in handling continuous actions has been extensively studied in the realm of robot learning \cite{mankowitz2019robust,perrusquia2021continuous,rahul2024deep}. A hallmark of a proficient path-planning algorithm lies in its ability to generate a minimal set of stable output points. Contrarily, algorithms such as the one proposed in \cite{ze} exhibit instability, rendering them impractical for real-world deployment. To address this, we employ two learning methods for the agent \cite{frameskip,caps}, to produce a model characterized by desired features such as action stability and simplicity.

Our methodology establishes a conducive learning environment for the path-planning task, anchored on three core pillars: a reward system centered on agent actions, action stability, and simplicity in action selection. Our results yield path-planning agents capable of generating intuitive trajectories for human comprehension. While the motion control aspect for SSL robots has been extensively explored in prior works \cite{araujo2023robocin,abiyev2017fuzzy,abiyev2018control}, our deliberate abstraction of this module enables our agents to seamlessly integrate into real-world setups, facilitating a plug-and-play deployment paradigm.

%% file: content/approach.tex
\section{Motion Planning Learning Environments}\label{sec:approach}

We adopted a similar environment structure guided by the fundamental principles of \cite{ze}. However, we explore RL-based path-planning to abstract the motion control module, following the pipeline presented in \fref{fig:general-workflow} \footnote{Agents, both trainable and trained, are available at \url{https://github.com/goncamateus/Planning-the-path-with-rl}}.

\subsection{Baseline Environment}
\label{subsec:approach_baseline}
Cruz \textit{et al.} \cite{ze} explore two RL-based motion planning approaches: RL-based path-planning to a given control system and a single-component approach, where the RL agent learns both motion planning and control. As a starting point for this work, we have successfully replicated their environments within the rSoccer framework \cite{rsoccer}, employing the first approach as a Baseline environment for this research.
The state space within these environments is characterized by a 13-dimensional vector: $(x^t$, $y^t$, $cos(\theta^t)$, $sin(\theta^t)$, $v_x^t$, $v_y^t$, $x^r$, $y^r$, $cos(\theta^r)$, $sin(\theta^r)$, $v_x^r$, $v_y^r$, $\omega^r)$, where $t$ and $r$ labels account for target and robot respectively. Also, in applying RL-based path-planning to a given motion control, the action space consists of a 6-tuple $(x, y, v_x, v_y, \sin\theta, \cos\theta)$, generating a sub-goal at each environment step. Notably, these vectors are subject to normalization concerning the maximum values in each dimension, except for the angular components, represented as sine and cosine values.

The reward system in \cite{ze} is defined by $R_T(s,a) = R_d(s, a) + R_\theta(s, a) + R_t(s, a)$, where:
\begin{align*}
    R_d(s, a) &= \begin{cases}
        -\text{d}(s, a), & \text{if } \text{dist}(s, a) > D_{th}\\
        10, & \text{otherwise}
    \end{cases} \\
    R_\theta(s,a) &= \begin{cases}
        \frac{-\delta(s, a)}{\pi}, & \text{if } \delta(s, a) > \theta_{th}\\
        1, & \text{otherwise}
    \end{cases} \\
    R_t(s,a) &= \begin{cases}
        1000, & \text{if } \text{dist}(s, a) > D_{th} \text{ and } \delta(s, a) > \theta_{th}\\
        0, & \text{otherwise}
    \end{cases}
\end{align*}
Here, $s$ and $a$ are state and action, \(\text{d}(s, a)\) and \(\delta(s, a)\) are the distance and angle difference to the target, with thresholds \(D_{th}\) and \(\theta_{th}\), respectively.
As elucidated in Section \ref{sec:related}, these rewards do not depend directly on the actions taken but on the robot's actuators. See in \fref{fig:env_baseline} a graphical representation of the angle difference and the distance to the target.

\subsection{Proposed Environment}

Firstly, we simplified the learning problem by imposing two main constraints: the agent's actions velocity components magnitude is constrained to zero; the goal also has the target velocity magnitude constrained to zero. Therefore, the Local Planning, as seen in \fref{fig:general-workflow}, is in form of $x, y, \theta, 0, 0$. The state and action spaces remain the same as in the Baseline.

The reward system proposed by \cite{ze} accounts only for the robot's properties, underscoring the agent's responsibility to optimize its actions and achieve the desired objectives. We introduced an action-centric reward structure by adding action features to our reward system, emphasizing states where the robot is on the cusp of achieving favorable outcomes within each reward component. 

\begin{figure}[ht]
    \centering
    \begin{subfigure}{0.3\textwidth}
        \centering
        \includegraphics[width=\textwidth]{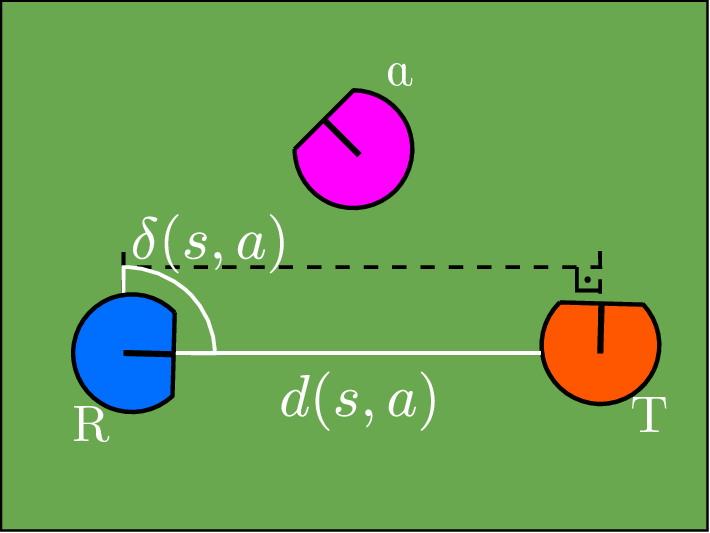}
        \caption{Baseline}
        \label{fig:env_baseline}
    \end{subfigure}
    \begin{subfigure}{0.3\textwidth}
        \centering
        \includegraphics[width=\textwidth]{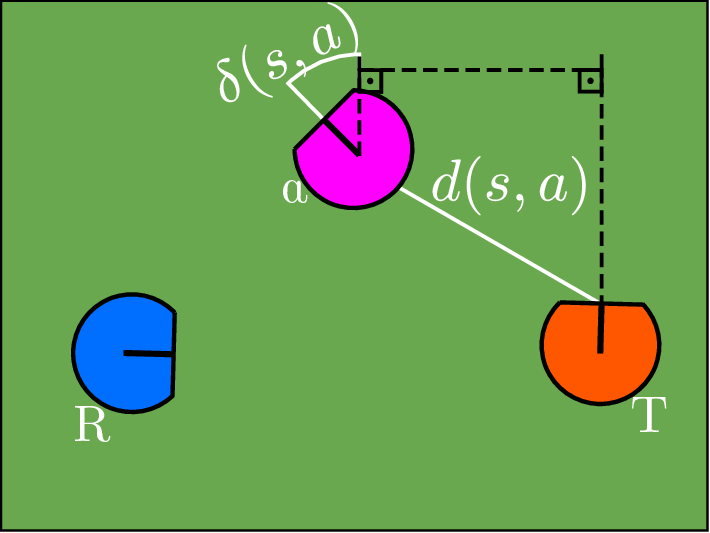}
        \caption{Proposed}
        \label{fig:env_proposed}
    \end{subfigure}
    \begin{subfigure}{0.3\textwidth}
        \centering
        \includegraphics[width=\textwidth]{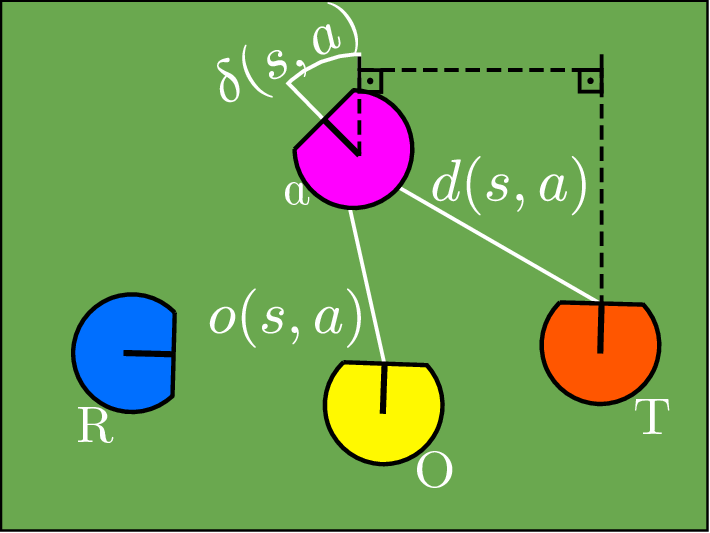}
        \caption{Obstacle}
        \label{fig:env_obstacle}
    \end{subfigure}
    \caption{Visualization of the ``Baseline", ``Proposed", and ``Obstacle" learning environments with example rewards. The circles blue (R), magenta (a), yellow (O), and orange (T) represent the robot's position, the action taken, the obstacle, and the target goal, respectively. The white arc and lines illustrate the angular difference, $\delta(s, a)$, and distances, $d(s, a)$, $o(s, a)$, to target and obstacle, respectively.}
\end{figure}

Therefore, the whole reward system, $R_d, R_\theta$, and $R_t$, outlined in Section \ref{subsec:approach_baseline} remains the same but changes the reference from the robot to the sub-goal action of our path-planning agent. Note in \fref{fig:env_proposed} a visual difference between our Proposed environment and the Baseline.

In our proposed reward system, the agent is penalized if the generated path is incompatible with the target. This strategic shift ensures that the agent's learned actions align more closely with the desired trajectory, fostering a genuine path-planning capability and reinforcing the agent's focus on optimal and effective target-reaching strategies.

\subsection{Proposed Obstacle Environment}
The obstacle setup inherits the Proposed environment and adds an enemy robot as the obstacle. Therefore, the state space for this environment is incremented by a 5-dimensional vector referent from the obstacle: $x^o, y^o, v_x^o, v_y^o, \omega^o$. The obstacle moves randomly, varying its speed in a range of $[0, 1] m/s$ each step.
In this environment, we also add penalties for hitting the obstacle and the actions being close to it. When the agent hits an obstacle, the episode fails, and the penalty is the inverse of reaching the target. 

At each step, we penalize the agent's proximity from the obstacle with a parametric Gaussian in function of the distance between the action and the obstacle $o(s,a)$, expected value $\mu=0$, and variation $\sigma=1$. Therefore, we can describe the reward system as $R_T(s,a)= R_{d}(s, a) + R_{\theta}(s, a) + R_{t}(s, a) + R_{obst}(s, a) + R_{hit}(s, a)$, where $R_{hit}(s, a) = -1000$ if the robot hits the obstacle; otherwise, $0$. See the addition of the obstacle distance reward in \fref{fig:env_obstacle}.



%% file: content/stability.tex
\section{Mitigating Action Instability}
\label{sec:stability}
The prominent issue of erratic and non-smooth actions learned by RL policies is well-recognized, posing potential concerns such as overheating and energy wastage problems in practical applications \cite{benchmarkingRLRW,benchmarkingRLCC}. In our context, this problem manifests as the agent controlling the robot not through angular or linear velocities but via poses. We introduce two techniques known to be effective in addressing this challenge in our training process.

The \textit{Frame skip} wrapper simplifies action sequences by repeating a given action for a specified number of steps. As it demonstrated to enhance agent performance in Atari games using Deep Q-Networks \cite{dqn,frameskip}, we apply this wrapper with a skip rate of $16$ in the rSoccer framework \cite{rsoccer}. This adjustment significantly reduces the number of actions per episode from $1200$ to a maximum of $75$.

Mysore \textit{et al.} \cite{caps} introduce an intuitive regularization technique, achieving an impressive $80\%$ reduction in power consumption during robot experiments. This technique, termed \textit{Conditioning for Action Policy Smoothness} (\textbf{CAPS}), introduces an additional loss term for the Actor in an Actor-Critic setup. CAPS aims to penalize actions that exhibit significant differences for similar states, promoting smoother and more controlled behavior.

%% file: content/results.tex
\section{Results} \label{sec:results}

We evaluate the agent's performance in the presented environments, namely Baseline, Proposed, and Obstacle, comparing their episodic lengths and the Cumulative Pairwise Action Distance (\textbf{CPAD}), a novel proposed measure designed to assess the stability of executed actions. This metric quantifies the overall displacement solely in the $(x, y)$ position and is mathematically defined as: $D(A) = \sum_{i=0}^{K-1} \text{Dist}(A_i, A_{i+1})$, where $A$ represents the actions undertaken throughout the episode, and $K$ denotes the episode length. 

CPAD offers valuable insights into the agent's behavioral stability. Specifically, a consistently directed focus on a particular $(x, y)$ position results in a cumulative distance of $0$. Conversely, rapid and erratic movements yield higher cumulative distances. As such, a lower CPAD is a crucial indicator of more stable and consistent actions, a pivotal attribute in motion planning.

Next topics unveils the results obtained from trained agents using the Soft Actor-Critic (\textbf{SAC}) \cite{sac} method in the setups described in Section \ref{sec:approach}. It is important to note that, given our use of the SAC algorithm for training agents, adjustments were made to accommodate the CAPS algorithm. 
SAC's policy entropy factor $\alpha$ is a parameter of the policy learning method to induce exploration during training. This parameter is originally set as trainable to, for example, explore at the beginning of training and then having stable actions. When applying CAPS at SAC, the $\alpha$ parameter hinds the stability loss, and the learning in general. Therefore, we set $\alpha$ as a constant throughout the SAC-CAPS algorithm execution.

\input{content/results/no-obstacle}
\input{content/results/obstacle}
\input{content/results/real}

%% file: content/results/no-obstacle.tex
\subsection{Comparing Baseline and Proposed Environments}
\label{subsec:res_noobs}
This evaluation assesses SAC agents in both Baseline and Proposed environments using four different setups: \textbf{1)} Vanilla, \textbf{2)} with Frame Skip, \textbf{3)} with CAPS, and \textbf{4)} with both Frame Skip and CAPS (FSCAPS). These setups provide insights into the model's adaptability within a real-world environment.

Table \ref{tab:results_no_obs} shows the proposed metrics results for obstacle-free environments. Notably, both metrics gradually decrease with the addition of action-stabilizing methods. The Proposed environment consistently outperforms the Baseline, showcasing improved agent motion.
\begin{table}[ht]
\caption{Test results distributions are presented for obstacle-free environments over 1000 episodes. Pairwise comparisons are conducted for each setup using the methods outlined in Section \ref{sec:stability}. The Vanilla experiments involve running agents in the environments described in Section \ref{sec:approach} without any action stabilizing method. The Frame Skip and CAPS experiments are similar but include either the frame skip or CAPS method. The FSCAPS experiments encompass both stabilization methods.}
\label{tab:results_no_obs}
\centering
\begin{tabular}{cc|ccc|ccc|}
\cline{3-8}
                                                   &             & \multicolumn{3}{c|}{Episode Length (steps)}                                                                                                   & \multicolumn{3}{c|}{CPAD (m)}                                                                                                                       \\ \cline{2-8} 
\multicolumn{1}{c|}{}                              & Environment & \multicolumn{1}{c|}{IQR}                                     & \multicolumn{1}{c|}{Max}                          & Min                        & \multicolumn{1}{c|}{IQR}                                        & \multicolumn{1}{c|}{Max}                           & Min                          \\ \hline
\multicolumn{1}{|c|}{}                             & Baseline    & \multicolumn{1}{c|}{1200 (1200 - 1200)}                      & \multicolumn{1}{c|}{1200}                         & 52                         & \multicolumn{1}{c|}{20.84 (14.42 - 26.75)}                      & \multicolumn{1}{c|}{47.66}                         & 1.39                         \\
\multicolumn{1}{|c|}{\multirow{-2}{*}{Vanilla}}       & Proposed    & \multicolumn{1}{c|}{\cellcolor[HTML]{FFFF00}\textbf{128 (102 - 152)}} & \multicolumn{1}{c|}{\cellcolor[HTML]{FFFF00}\textbf{1200}} & \cellcolor[HTML]{FFFF00}\textbf{35} & \multicolumn{1}{c|}{\cellcolor[HTML]{FFFF00}\textbf{6.03 (3.93 - 8.07)}} & \multicolumn{1}{c|}{\cellcolor[HTML]{FFFF00}\textbf{16.43}} & \cellcolor[HTML]{FFFF00}\textbf{0.20} \\ \hline
\multicolumn{1}{|c|}{}                             & Baseline    & \multicolumn{1}{c|}{198 (150 - 265)}                         & \multicolumn{1}{c|}{1200}                         & 49                         & \multicolumn{1}{c|}{13.53 (11.03 - 16.55)}                      & \multicolumn{1}{c|}{38.40}                         & 3.46                         \\
\multicolumn{1}{|c|}{\multirow{-2}{*}{Frame Skip}} & Proposed    & \multicolumn{1}{c|}{\cellcolor[HTML]{FFFF00}\textbf{119 (95 - 141)}}  & \multicolumn{1}{c|}{\cellcolor[HTML]{FFFF00}\textbf{223}}  & \cellcolor[HTML]{FFFF00}\textbf{33} & \multicolumn{1}{c|}{\cellcolor[HTML]{FFFF00}\textbf{1.31 (0.60 - 2.12)}} & \multicolumn{1}{c|}{\cellcolor[HTML]{FFFF00}\textbf{6.80}}  & \cellcolor[HTML]{FFFF00}\textbf{0.01} \\ \hline
\multicolumn{1}{|c|}{}                             & Baseline    & \multicolumn{1}{c|}{1200 (731 - 1200)}                       & \multicolumn{1}{c|}{1200}                         & 37                         & \multicolumn{1}{c|}{13.97 (9.61 - 16.31)}                       & \multicolumn{1}{c|}{22.86}                         & 1.91                         \\
\multicolumn{1}{|c|}{\multirow{-2}{*}{CAPS}}       & Proposed    & \multicolumn{1}{c|}{\cellcolor[HTML]{FFFF00}\textbf{113 (91 - 134)}}  & \multicolumn{1}{c|}{\cellcolor[HTML]{FFFF00}\textbf{1200}} & \cellcolor[HTML]{FFFF00}\textbf{33} & \multicolumn{1}{c|}{\cellcolor[HTML]{FFFF00}\textbf{1.29 (0.81 - 1.87)}} & \multicolumn{1}{c|}{\cellcolor[HTML]{FFFF00}\textbf{3.81}}  & \cellcolor[HTML]{FFFF00}\textbf{0.14} \\ \hline
\multicolumn{1}{|c|}{}                             & Baseline    & \multicolumn{1}{c|}{\cellcolor[HTML]{FFFF00}\textbf{109 (87 - 130)}}  & \multicolumn{1}{c|}{\cellcolor[HTML]{FFFF00}\textbf{196}}  & \cellcolor[HTML]{FFFF00}\textbf{34} & \multicolumn{1}{c|}{0.58 (0.36 - 0.87)}                         & \multicolumn{1}{c|}{2.22}                          & 0.02                         \\
\multicolumn{1}{|c|}{\multirow{-2}{*}{FSCAPS}}     & Proposed    & \multicolumn{1}{c|}{114 (92 - 137)}                          & \multicolumn{1}{c|}{193}                          & 31                         & \multicolumn{1}{c|}{\cellcolor[HTML]{FFFF00}\textbf{0.06 (0.05 - 0.08)}} & \multicolumn{1}{c|}{\cellcolor[HTML]{FFFF00}\textbf{0.16}}  & \cellcolor[HTML]{FFFF00}\textbf{0.00} \\ \hline
\end{tabular}
\end{table}


When evaluating individual experiments through a ``Baseline \textit{vs} Proposed'' lens, our proposed environment consistently demonstrates superior performance in both episode length and CPAD metrics. The Vanilla experiment, notable improvements of \textbf{90\% for episode length} and \textbf{71\% for CPAD} were observed. Incorporating the Frame Skip wrapper yielded a significant performance enhancement, reducing task completion time by approximately \textbf{80 steps} (equivalent to 6\% of the total episode) and achieving a \textbf{90\%} improvement in the CPAD metric. In the CAPS experiment, inspired by \cite{caps}, we achieved a \textbf{78\%} CPAD reduction compared to Vanilla-Proposed. This outcome underscores the limitations of baseline-trained agents in attaining the precision of our proposed approach.

Combining action-stabilizing methods in the FSCAPS experiments showcases enhanced performance. The Baseline-trained agent exhibits marginally improved episode length, while the Proposed-trained agent demonstrates a remarkable \textbf{tenfold improvement} in action precision. Therefore, the gain in episode length is deemed negligible compared to the Proposed agent's superior transfer capability to a real-world environment.

\begin{figure}[ht]
    \centering
    \begin{subfigure}{0.24\textwidth}
        \centering
        \includegraphics[width=\textwidth]{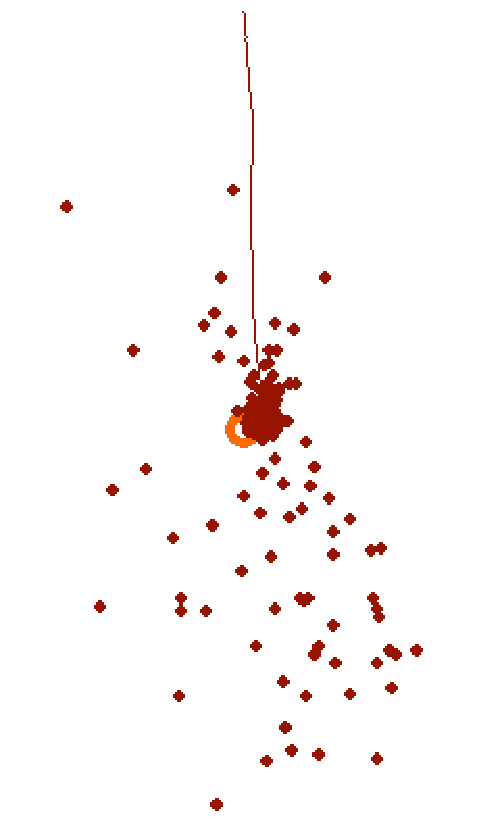}
        \subcaption{Vanilla}
        \label{fig:res_Vanilla}
    \end{subfigure}
    \begin{subfigure}{0.24\textwidth}
        \centering
        \includegraphics[width=\textwidth]{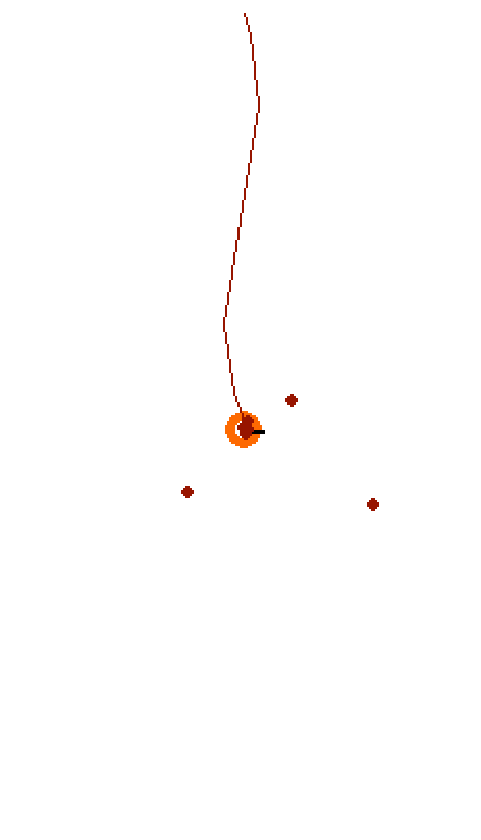}
        \subcaption{Frame Skip}
        \label{fig:res_fs}
    \end{subfigure}
    \begin{subfigure}{0.24\textwidth}
        \centering
        \includegraphics[width=\textwidth]{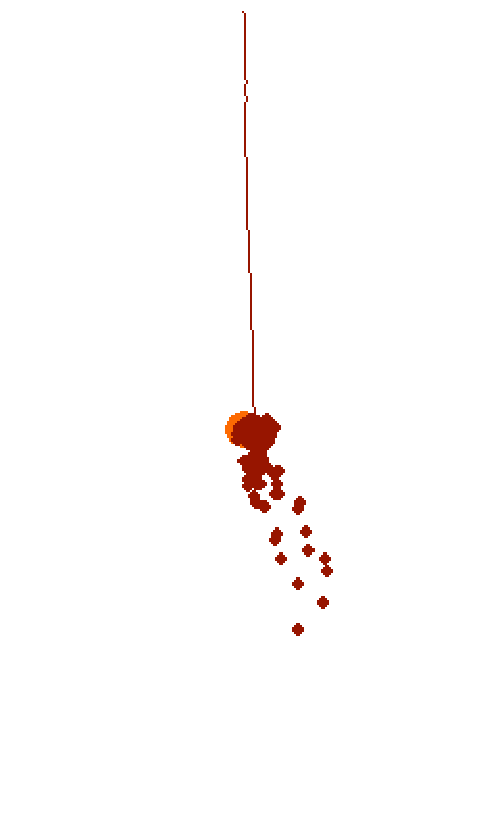}
        \subcaption{CAPS}
        \label{fig:res_caps}
    \end{subfigure}
    \begin{subfigure}{0.24\textwidth}
        \centering
        \includegraphics[width=\textwidth]{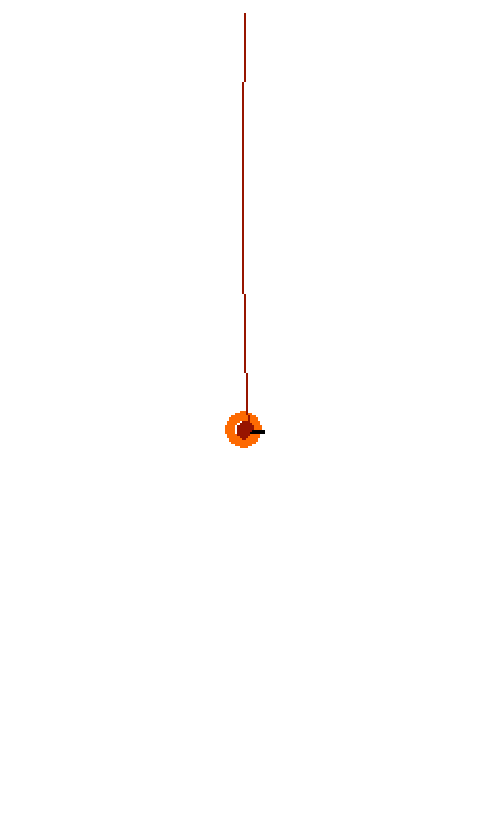}
        \subcaption{FSCAPS}
        \label{fig:res_fscaps}
    \end{subfigure}
    \caption{Trained Soft Actor-Critic agents operating within the Proposed environment across various experiments. The orange circle denotes the target goal, while the red dots trace the actions executed by the agent throughout the episode. The red line visually maps the trajectory followed by the robot. As depicted in \fref{fig:res_fscaps}, the agent demonstrates precision by consistently reaching the target without deviation.}
    \label{fig:results}
\end{figure}

In comparing experiments within our proposed environment, the final temporal gain between Vanilla and FSCAPS experiments is modest, with only 14 simulated steps (1\% of the total episode) faster. However, examining the CPAD metric reveals a substantial increase in action stability with each addition, resulting in a final gain \textbf{100 times better} than the environment alone. See in \fref{fig:results} trained agents for our proposed environment in each experiment. These compelling results encourage further research, particularly in environments with obstacle setups.

%% file: content/results/obstacle.tex
\subsection{Evaluating Obstacle Avoidance}
\label{subsec:res_obs}
The proposed Obstacle environment is evaluated using SAC agents following the scheme outlined in Section \ref{subsec:res_noobs}. This environment features a simple scenario with a single mobile robot acting as the obstacle. Uniquely in this setup, the percentage of hitting the obstacle is introduced as an additional metric, providing insights into the adaptability of our previous research in obstacle-laden environments.

Table \ref{tab:results_obs} presents results from experiments in the Obstacle environment. Preliminary results indicate that the baseline method struggled to solve the task consistently within the imposed training time. In contrast, our final method, FSCAPS, introduces a robust and adaptable approach for environments with obstacle. Understandably, variations in actions are observed as the agent needs to navigate around the moving obstacle. The low rate of collisions suggests that the agent effectively learned the purpose of the environment, even at maximum speed, with the episode length nearly identical to the setup without obstacles.


\begin{table}[htb]
\caption{Test results distributions are presented for the environment with obstacle over 1000 episodes.}
\label{tab:results_obs}
\centering
\begin{tabular}{c|ccc|ccc|c|}
\cline{2-8}
                                 & \multicolumn{3}{c|}{Episode Length (steps)}                                                                                                                              & \multicolumn{3}{c|}{CPAD (m)}                                                                                                                                                 & Collision                            \\ \cline{2-8} 
                                 & \multicolumn{1}{c|}{IQR}                                              & \multicolumn{1}{c|}{Max}                                   & Min                                 & \multicolumn{1}{c|}{IQR}                                                 & \multicolumn{1}{c|}{Max}                                   & Min                                   &                                         \\ \hline
\multicolumn{1}{|c|}{Vanilla}       & \multicolumn{1}{c|}{1200 (1200 - 1200)}                               & \multicolumn{1}{c|}{1200}                                  & 132                                 & \multicolumn{1}{c|}{12.38 (10.70 - 14.21)}                               & \multicolumn{1}{c|}{35.03}                                 & 4.10                                  & 12.38\%                                 \\ \hline
\multicolumn{1}{|c|}{Frame Skip} & \multicolumn{1}{c|}{1200 (1200 - 1200)}                               & \multicolumn{1}{c|}{1200}                                  & 148                                 & \multicolumn{1}{c|}{6.03 (5.24 - 6.90)}                                  & \multicolumn{1}{c|}{11.35}                                 & 2.23                                  & 5.61\%                                  \\ \hline
\multicolumn{1}{|c|}{CAPS}       & \multicolumn{1}{c|}{1200 (1200 - 1200)}                               & \multicolumn{1}{c|}{1200}                                  & 670                                 & \multicolumn{1}{c|}{11.35 (10.55 - 12.17)}                               & \multicolumn{1}{c|}{20.60}                                 & 4.10                                  & 1.32\%                                  \\ \hline
\multicolumn{1}{|c|}{FSCAPS}     & \multicolumn{1}{c|}{\cellcolor[HTML]{FFFF00}\textbf{115 (113 - 158)}} & \multicolumn{1}{c|}{\cellcolor[HTML]{FFFF00}\textbf{1200}} & \cellcolor[HTML]{FFFF00}\textbf{79} & \multicolumn{1}{c|}{\cellcolor[HTML]{FFFF00}\textbf{1.27 (1.06 - 1.53)}} & \multicolumn{1}{c|}{\cellcolor[HTML]{FFFF00}\textbf{5.87}} & \cellcolor[HTML]{FFFF00}\textbf{0.55} & \cellcolor[HTML]{FFFF00}\textbf{0.61\%} \\ \hline
\end{tabular}
\end{table}

%% file: content/results/real.tex
\subsection{Real environment results}
\label{subsec:res_real}

After obtaining stable and accurate results in a simulated environment, the trained agents were subsequently deployed in a real-world setting. We integrated our model into the RobôCIn SSL-Unification software \cite{araujo2023robocin} using the TorchScript module \cite{paszke2019pytorch} for compilation and adaptation purposes. It is important to note that the RobôCIn SSL-Unification software employs its own motion control system, whereas our agents were trained using a general omnidirectional motion control approach \cite{siegwart2011introduction}.

The agents utilized in these experiments included the Vanilla agent, trained in an obstacle-free environment, and the FSCAPS agent, trained in our proposed ``Obstacle'' environment. For experiments involving obstacles, we employed a fixed obstacle to ensure safety.

Table \ref{tab:results_real} presents the episode length and CPAD for the experiments on the real-world. The Vanilla agent failed to complete the task in any of the ten trials. The CPAD metric suggests unstable paths extending to a length of 25 fields, turning it unpractical to real-world.

In contrast, the FSCAPS agent exhibited consistent results across both environment setups. These findings align with our approach of developing a methodology that is agnostic to the motion control system, thereby facilitating deployment in real-world scenarios.

\begin{table}[]
\centering
\caption{Test results distributions are presented for real-world setup over 10 episodes.}
\begin{tabular}{c|ccc|ccc|}
\cline{2-7}
                               & \multicolumn{3}{c|}{Episode Length (steps)}                                & \multicolumn{3}{c|}{CPAD (m)}                                                        \\ \cline{2-7} 
                               & \multicolumn{1}{c|}{IQR}                & \multicolumn{1}{c|}{Max}  & Min  & \multicolumn{1}{c|}{IQR}                      & \multicolumn{1}{c|}{Max}    & Min    \\ \hline
\multicolumn{1}{|c|}{Vanilla}  & \multicolumn{1}{c|}{1200 (1200 - 1200)} & \multicolumn{1}{c|}{1200} & 1200 & \multicolumn{1}{c|}{227.93 (227.06 - 230.79)} & \multicolumn{1}{c|}{235.06} & 225.45 \\ \hline
\multicolumn{1}{|c|}{FSCAPS}   & \multicolumn{1}{c|}{159 (155 - 160)}    & \multicolumn{1}{c|}{162}  & 135  & \multicolumn{1}{c|}{0.10 (0.10 - 0.10)}       & \multicolumn{1}{c|}{0.99}   & 0.09   \\ \hline
\multicolumn{1}{|c|}{Obstacle} & \multicolumn{1}{c|}{140 (131 - 151)}    & \multicolumn{1}{c|}{171}  & 109  & \multicolumn{1}{c|}{2.10 (2.04 - 2.15)}       & \multicolumn{1}{c|}{2.32}   & 2.01   \\ \hline
\end{tabular}
\label{tab:results_real}
\end{table}

%% file: content/discussion.tex
\section{Discussion}
\label{sec:discussion}

In analyzing agent adaptability, FSCAPS demonstrates significant superiority in acquiring efficient strategies in challenging environments. The combination of Frame Skip with CAPS accelerates adaptation to obstacle dynamics, evidenced by reduced episode durations and collision rates. This suggests more effective learning and better generalization compared to isolated methods. Environmental parameters significantly impact path planning efficacy, with FSCAPS exhibiting greater robustness against environmental variations. This emphasizes the need for strategies resilient to fluctuations.

Compared to state-of-the-art approaches, FSCAPS not only outperforms in efficiency and safety but also strikes a balanced solution, especially in complex obstacle environments where real-time adaptation is crucial. Navigation failures revealed that Pure and Frame Skip methods struggled with unpredictable obstacles. The integration of CAPS enhances predictability and action smoothness, reducing collision rates.

While longer episodes offer more experience, they can induce fatigue. FSCAPS reduces episode duration without compromising safety, offering a sustainable training approach. However, generalizing FSCAPS to untested scenarios requires further exploration to validate its real-world robotic applications.

%% file: content/conclusion.tex
\section{Conclusion} \label{sec:clonclusion}
In this study, we comprehensively evaluated Reinforcement Learning (RL) applied to the intricate challenges of robot motion planning within the RoboCup Small Size League (SSL). Leveraging a heuristic control approach, we explored various learning environments, comparing the performance of SAC agents in both Baseline and Proposed setups.

Our exploration in obstacle-free environments showcased the Proposed environment's effectiveness in reducing action instability, and enhancing agent performance. The Frame Skip wrapper further improved efficiency, and the combined use of Frame Skip and CAPS in the ``FSCAPS" experiments demonstrated significant performance gains.

Transitioning to obstacle-laden environments, the initial experiments faced challenges in consistently solving tasks, emphasizing the complexity introduced by mobile obstacles. However, our final method, ``FSCAPS", exhibited evident adaptability, effectively navigating around obstacles with a minimal rate of collision. 

Furthermore, we validated the adaptability of our model in a real-world setting without compromising performance in terms of trajectory length or precision. This corroborates the efficacy of our approach, emphasizing the importance of initially addressing high-level tasks for seamless model adaptation between simulation and real-world deployment.

These results underscore the significance of our research in advancing RL techniques for robot motion planning, particularly in dynamic and challenging SSL environments. The strategic integration of heuristic control and innovative stabilizing techniques demonstrates the potential for improved decision-making and adaptability in dynamic robotic scenarios, paving the way for advancements in the field of autonomous robotics.

